\documentclass{article}

\usepackage[final]{corl_2026} 
\usepackage{amsmath}
\usepackage{amssymb}
\usepackage{graphicx}
\usepackage{booktabs}
\usepackage{makecell}

\title{Incorporating World Models in Decision Making for Next Best View Selection in Active Sensing}

\author{
  Nikita Sarawgi\\
  GrayMatter Robotics Inc.\\
  \texttt{nikita.sarawgi@graymatter-robotics.com} \\
  \And
  Omey M. Manyar\\
  GrayMatter Robotics Inc.\\
  \texttt{omey.manyar@graymatter-robotics.com} \\
  \And
  Satyandra K. Gupta\\
  GrayMatter Robotics Inc.\\
  \texttt{skgupta@graymatter-robotics.com} \\
}
\begin{document}
\maketitle
\raggedbottom

\begin{abstract}
    Efficient robotic 3D reconstruction requires selecting views that contribute new surface measurements with few physical captures. Pretrained world models can guide this choice, but a predicted surface may not produce valid depth measurements under the operating conditions of the deployed sensor. We present a deployment adaptation methodology for next-best-view selection that learns a sensor-return adapter from limited local calibration data while keeping the pretrained predictor fixed. The adapter estimates measurement dropout over each predicted surface, allowing the selection method to favor views likely to fill gaps in the measured reconstruction. In a real-robot experiment with a structured-light sensor, the adapted method achieves higher measured surface coverage with 60\% fewer captures than nominal selection. The adapted method recovers a surface region that the nominal method repeatedly targets but fails to measure. These results suggest that robots can reuse pretrained predictive models while adapting their decisions to local operating conditions in data-scarce environments.
\end{abstract}



\section{Introduction}
\label{sec:introduction}
\vspace{-5pt}

Object reconstruction from sensor measurements supports several crucial applications such as robotic inspection, manipulation, and surface processing~\cite{scott2003viewplanning,raible2023pathplanning}. In industrial processing which requires millimeter-level accuracy and high surface resolution, reconstruction quality is critical. Access to a CAD model or predefined geometric primitives cannot be assumed in high-mix production~\cite{leonard2026highmix}; even when available, nominal geometry alone does not capture the surface deviations or placement of the manufactured part~\cite{raible2023pathplanning}. Processing at this tolerance therefore requires measurements of the part as presented to the robot. Despite advances in structured-light and laser-line sensors, measurement quality remains sensitive to deployment conditions. In high-mix manufacturing, parts vary in geometry, texture, and finish~\citep{leonard2026highmix}, where sensor settings, illumination, and surface reflectivity dictate whether geometrically visible regions yield valid depth measurements, and missing regions can change with viewpoint~\citep{zhang2020shiny,yang2022nbvreflective}. Because successful strategies on one setup can fail on another, sensing must adapt to local deployment conditions.

Pretrained action-conditioned world models guide the selection of sensing actions by predicting state transitions before execution~\cite{agarwal2025cosmos,assran2025vjepa2}. However, selection using nominal predictions assumes visible surfaces equal valid depth measurements. When predictions do not account for local sensor-return behavior, selection risks overvaluing views that give invalid measurements, resulting in wasted captures. Fine-tuning on cell-specific data consumes costly production time, whereas discarding predictions ignores generalizable task dynamics. These constraints motivate preserving a fixed predictor for general task dynamics while learning sensor behavior from limited local experience.

Our main contribution is a methodology that adapts how predictions guide action selection under local conditions using limited calibration data while keeping the pretrained model fixed. 
We demonstrate this framework in next-best-view (NBV) selection, where a manipulator selects views to reconstruct a static object. Traditional NBV methods prioritize geometric visibility and surface coverage~\cite{isler2016information,delmerico2018comparison,peralta2020next,guedon2022scone,chen2024gennbv}. Our framework accounts for whether the deployed sensor obtains valid depth from visible surfaces. A fixed action-conditioned world model provides candidate views and expected surface geometry~\cite{wu2026framecrafter} while a separate adapter predicts measurement dropout under deployment conditions.
The view-selection method filters predicted geometry using these estimates before selection. Decoupling task-level prediction from local sensor behavior ensures calibration data is spent on learning deployment-specific sensor returns, enabling model reuse across setups.

We evaluate this framework on a real robot, training the adapter on 50 captures across four helmets and testing on a held-out reflective helmet under fixed lighting and sensor settings. While nominal selection repeatedly selects views that produce invalid depth, realizing 11.7\% of its predicted coverage gain, deployment-aware selection filters candidates by estimated sensor return. It reaches the 98\% coverage target in four captures, whereas nominal method fails after ten. These results show that a lightweight adapter prevents repeated acquisition failures using fixed pretrained predictions.

\section{Problem Definition}
\label{sec:probdef}
\vspace{-5pt}
We consider active sensing with a manipulator $M$ equipped with a sensor $S$ to reconstruct a static object $O$ with unknown surface $\mathcal{S}$ within a known bounding box $\mathcal{W} \subset \mathbb{R}^3$, next-best-view (NBV) selection~\citep{isler2016information,delmerico2018comparison}, given a fixed pretrained action-conditioned world model $F_\theta$ and limited local data $\mathcal{D}_\eta$ to characterize domain-specific sensor-return behavior.

Initial observations provide context $\mathcal{C}_0 = \{(I_i, T_i)\}_{i=1}^{m}$ of $m$ RGB images and camera poses. For candidate pose $T$, $F_\theta$ uses $\mathcal{C}_0$ to estimate the expected visible surface $\hat{\mathcal{P}}(T)$. At step $t$, policy $\pi$ selects action $a_t = T \in \mathcal{V}$ from information state $s_t = (\mathcal{C}_0, \mathcal{P}_t)$, where $\mathcal{P}_t$ holds measurements gathered so far, and $\mathcal{V} \subset \mathrm{SE}(3)$ contains feasible, collision-free poses.

A surface point $\mathbf{x} \in \mathcal{S}$ is visible from pose $T$ ($v(\mathbf{x}, T) = 1$) if unoccluded and within sensor's field of view and depth range. Visibility does not guarantee a valid depth measurement ($b = 1$); otherwise a dropout ($b = 0$). The conditional return probability
\begin{equation}
r_\eta(\mathbf{x}, T) = \Pr(b = 1 \mid v = 1, \mathbf{x}, T, \eta)
\end{equation}
depends on deployment conditions $\eta$, which encapsulate the sensor, its settings, ambient illumination, and surface properties~\citep{zhang2020shiny,yang2022nbvreflective,maken2025improving}. We estimate $r_\eta$ from limited local data $\mathcal{D}_\eta$ and score predicted surfaces $\hat{\mathcal{P}}(T)$, with $F_\theta$ fixed. For voxel side length $h$, coverage $C_h(\mathcal{P}_t)$ is the fraction of surface-intersecting voxels containing at least one valid measurement. We minimize expected captures to reach $C_h(\mathcal{P}_t) \geq \rho$, $\rho \in (0, 1]$, reducing overall acquisition time. This excludes context ($\mathcal{C}_0$) and calibration captures. Ground-truth surface $\mathcal{S}$ is used for evaluation and is inaccessible to method.


\section{Methodology}
\label{sec:methodology}
\vspace{-5pt}

\begin{figure}[t]
    \centering
    \includegraphics[width=0.92\textwidth]{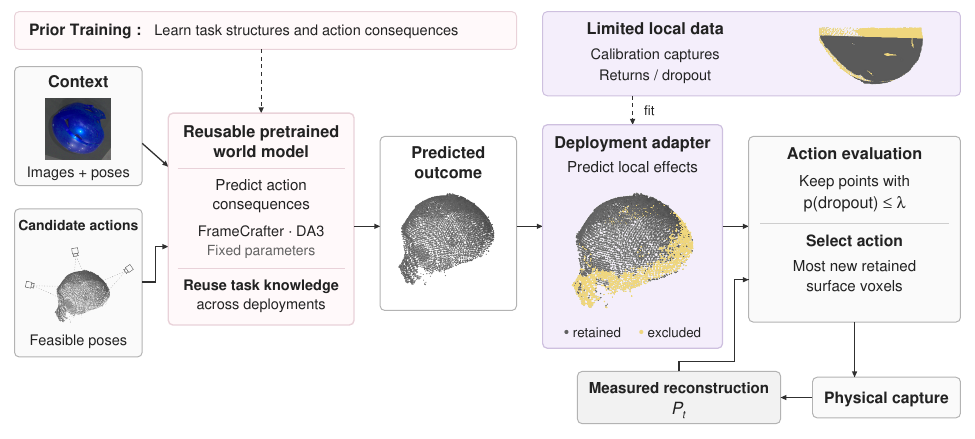}
    \caption{Deployment-aware next-best-view selection. A fixed predictor supplies geometry; a locally learned adapter estimates sensor return for action scoring. Captures update the reconstruction.}
    \label{fig:overview}
\end{figure}

Our focus is deployment adaptation, not world-model design. In static-scene
sensing, we use a task-specific world model to predict the scene as viewed
after a candidate sensing action. FrameCrafter supplies action-conditioned
appearance and Depth Anything~3 infers geometry; together they implement
this predictive role. We keep both fixed and learn sensor-return behavior
locally (Figure~\ref{fig:overview}), assuming their geometry remains useful.
The framework can instead use another pretrained or custom predictor
that supplies compatible candidate appearance and geometry estimates.

\noindent\textbf{Reusable surface prediction.}
We use FrameCrafter~\citep{framecrafter} to predict visual observations $\hat{I}(T)$ from camera views using a sparse, unordered set of context images $\mathcal{C}_0$. Given a candidate pose $T$:
\begin{equation}
\hat{I}(T) = F_\theta(\mathcal{C}_0, T).
\label{eq:framecrafter}
\end{equation}
Images alone cannot be compared with measured surface coverage. We use Depth Anything~3 (DA3)~\citep{da3} to jointly estimate geometry using the context images and $\hat{I}(T)$, giving the world-frame point cloud $P(T)$ visible from $T$. Predicted points $P(T)$ are used only for scoring candidate views; the active reconstruction $P_t$ contains strictly physical sensor measurements.

\noindent\textbf{Deployment adapter.}
The adapter estimates return probability $r_\eta$ from local features $\mathbf{z}=(d,\alpha,i)$: viewing distance $d$, incidence angle $\alpha$~\citep{maken2025improving}, and normalized intensity $i$. These features capture sensing geometry and appearance without requiring global object context. Training samples $\mathcal{D}_\eta=\{(\mathbf{z}_j,\delta_j)\}_{j=1}^{n}$ are generated by comparing calibration captures against a dense reference surface $\mathcal{S}_{\mathrm{ref}}$. Returned pixels are labeled $b=1$ and missing pixels $b=0$, producing dropout labels $\delta=1-b$ where $\mathcal{S}_{\mathrm{ref}}$ or adjacent returns confirm surface presence; occlusions and out-of-range pixels are excluded ($v=1$). A sparse variational Gaussian process classifier~\citep{hensman2015scalable} with parameters $\phi$ predicts dropout probability $p_{\phi,\eta}(\mathbf{z}) \approx \Pr(\delta=1 \mid \mathbf{z},\mathcal{D}_\eta)$. The return probability is estimated as $r_\eta(\mathbf{x},T) = 1 - p_{\phi,\eta}(\mathbf{z}(\mathbf{x},T))$. During selection, $d$ and $\alpha$ are computed from $\hat{\mathcal{P}}(T)$, and $i$ is taken from $\hat{I}(T)$. The adapter is trained offline and remains fixed.

\noindent\textbf{Acquisition score.}
For candidate poses $T \in \mathcal{V}_c \subset \mathcal{V}$, we retain visible predicted points $\hat{\mathcal{P}}(T)$ inside bounding box $\mathcal{W}$ satisfying $p_{\phi,\eta}(\mathbf{z}(\mathbf{x},T)) \leq \lambda$, guaranteeing return probability $\ge 1-\lambda$ ($\lambda=1$ recovers nominal selection). Using the voxel grid of Section~\ref{sec:probdef}, the greedy method selects the pose maximizing the count of unmeasured voxels containing retained points. Context $\mathcal{C}_0$ and model parameters remain fixed.


\section{Experiments}
\label{sec:experiments}
\vspace{-5pt}

We evaluate whether a separate deployment adapter learned from limited
local calibration data improves action selection while reusing a fixed
pretrained world model to predict action consequences.
We compare next-best-view selection with and without the adapter's
sensor-return estimates.

\begin{figure}[t]
    \centering
    \includegraphics[width=0.82\linewidth]{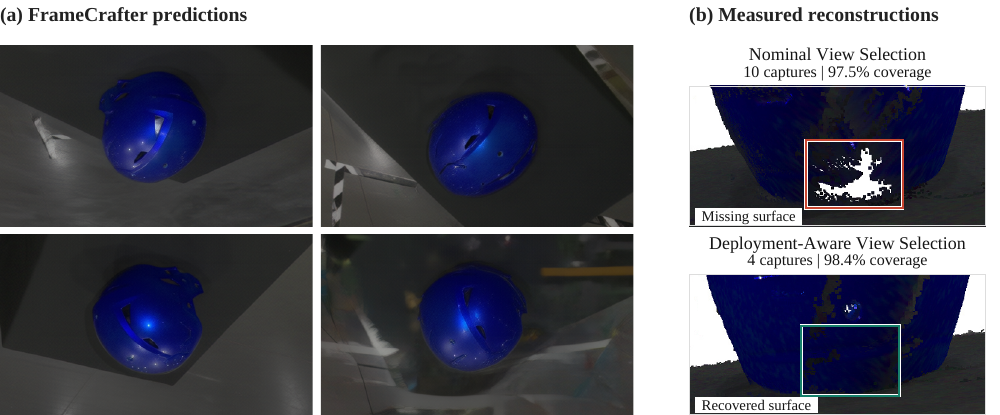}
    \caption{(a) FrameCrafter RGB predictions at 4 candidate views of the held-out blue helmet. (b) Measured reconstructions: nominal selection leaves a lower-edge gap after 10 captures (97.5\% coverage), while deployment-aware selection recovers it within 4 (98.4\%). Boxes highlight this region.}
    \label{fig:surface-reconstruction}
\end{figure}

\noindent\textbf{Experimental Setup.}
We use a FANUC M-710iC robot with a Zivid~2+ LR110 structured-light sensor~\citep{zivid2plus} to scan five reflective helmets of varying size, surface pattern, and color. We train the adapter on $10^6$ uniformly sampled pixels from 50 captures of four helmets, using dense reference surfaces assembled from these captures for dropout labels, and evaluate it on the held-out blue helmet. The initial context $\mathcal{C}_0$ consists of $m=6$ posed RGB images, which FrameCrafter uses to predict RGB observations at candidate viewpoints (Figure~\ref{fig:surface-reconstruction}(a)). Candidate camera poses are sampled on a sphere centered on the helmet, oriented toward its center, and retained after feasibility checks. The robot configuration, sensor settings, and lighting are fixed across calibration and evaluation. For this experiment, we use depth projections from the reference scan, as the current predicted depth requires further alignment to prevent small geometric offsets from inflating voxel-based coverage gains.

\noindent\textbf{View-selection comparison and evaluation.}
We compare \emph{nominal selection}, which scores predicted new surface coverage without dropout filtering ($\lambda=1$), with \emph{deployment-aware selection}, which excludes predicted points with dropout probability above $20\%$ ($\lambda=0.2$). Both use the same context images, candidate poses, and acquisition rule; FrameCrafter and Depth Anything~3 remain fixed. Coverage is computed from measured points on the voxel grid against the blue helmet's reference scan. We use a $98\%$ target to allow for possible artifacts in the reference scan, such as residual layering, that can populate voxels not intersecting the physical surface. Each method is evaluated once from $\mathcal{P}_0=\emptyset$, stopping at this target or after 10 captures; counts exclude context acquisitions.

\begin{table}[t]
\centering
\caption{Reconstruction results on the held-out blue helmet.}
\label{tab:results}
\begin{tabular}{lccc}
\toprule
Method & \makecell{Final coverage\\(\%)} & \makecell{Number of Captures} & \makecell{Realized / predicted gain\\after capture 2 (\%)} \\
\midrule
Nominal Selection & 97.5 & 10 & 11.7 \\
Deployment-aware Selection & 98.4 & 4 & 114.8 \\
\bottomrule
\end{tabular}
\end{table}

\noindent\textbf{Effect on surface reconstruction.}
Deployment-aware selection reaches 98.4\% coverage of the held-out blue
helmet after four captures (Table~\ref{tab:results}). Nominal selection
reaches 95.8\% after four captures and 97.5\% after ten, remaining below
the 98\% target. The nominal reconstruction retains a substantial gap
near the helmet's lower edge after ten captures, whereas deployment-aware
method recovers this region (Figure~\ref{fig:surface-reconstruction}). Both methods select the same
pose for their second capture, reaching comparable coverage: 95.5\% for
nominal method and 95.1\% for deployment-aware method. Over subsequent
captures, however, nominal method without the adapter realizes only 11.7\% of its predicted
coverage gain, compared to 114.8\% with it.
Nominal selection repeatedly selects views whose predicted point
clouds include surface in the missing region, but physical captures from
these views yield no valid depth returns from that region. Because
the region remains unmeasured, its predicted
points continue to contribute to acquisition scores, prompting
further unsuccessful captures. The adapter filters the same
candidate predictions using the sensor-return estimates and
selects views that yield valid depth measurements from the missing region. The substantial gap leaves no measured surface geometry for planning tool contact here. Routing around it can require separate toolpath segments, not only increasing manufacturing cycle time and leaving the surface unprocessed.

\section{Conclusion}
\label{sec:conclusion}
\vspace{-5pt}
This work explores how robots can reuse pretrained predictive models across deployments while adapting to local conditions through limited experience. Sharing an adapter across parts and environments with similar sensing conditions could allow the same limited calibration data to support additional deployments. The present evaluation covers an object from a class of geometrically similar parts under fixed sensing conditions, and calibration depends on reference surfaces. Further work should establish when changes in materials, sensors, or lighting require adapter updates.

\clearpage


\bibliography{example}  

\end{document}